%% file: 0_main.tex
\documentclass[final]{cvpr}

\usepackage{times}
\usepackage{epsfig}
\usepackage{graphicx}
\usepackage{amsmath}
\usepackage{amssymb}

\usepackage{xcolor}
\usepackage{subfig}
\usepackage{caption}
\usepackage{amsmath}
\usepackage{amssymb}
\usepackage{booktabs}
\usepackage[switch]{lineno}

\newtheorem{lemma}{Lemma}

\newcommand{\bx}{\mathbf{x}}
\newcommand{\bz}{\mathbf{z}}
\newcommand{\be}{\mathbf{e}}
\newcommand{\hX}{\hat{X}}
\newcommand{\hbx}{\hat{\mathbf{x}}}
\newcommand{\PP}{\mathbb{P}}

\newcommand{\bD}{\mathbf{D}}
\newcommand{\bG}{\mathbf{G}}
\newcommand{\bE}{\mathbf{E}}
\newcommand{\cL}{\mathcal{L}}
\usepackage{booktabs}
\usepackage{multirow}

\usepackage[pagebackref=true,breaklinks=true,colorlinks,bookmarks=false]{hyperref}

\def\cvprPaperID{10147} 
\def\confYear{CVPR 2021}

\begin{document}

\title{Unsupervised Anomaly Detection with Adversarial Mirrored AutoEncoders}

\author{Gowthami Somepalli\textsuperscript{\rm 1}, 
Yexin Wu\textsuperscript{\rm 1}, 
Yogesh Balaji\textsuperscript{\rm 1}, 
Bhanukiran Vinzamuri\textsuperscript{\rm 2},
Soheil Feizi\textsuperscript{\rm 1}\\
\and 
\textsuperscript{\rm 1}University of Maryland, College Park\\
{\tt\small \{gowthami,yogesh,soheil\}@cs.umd.edu}\\
{\tt\small ywu12319@terpmail.umd.edu}
\and
\textsuperscript{\rm 2}IBM Research\\
{\tt\small bhanu.vinzamuri@ibm.com}
}

\maketitle

\begin{abstract}
Detecting out of distribution (OOD) samples is of paramount importance in all Machine Learning applications. Deep generative modeling has emerged as a dominant paradigm to model complex data distributions without labels. However, prior work has shown that generative models tend to assign higher likelihoods to OOD samples compared to the data distribution on which they were trained. First, we propose Adversarial Mirrored Autoencoder (AMA), a variant of Adversarial Autoencoder, which uses a mirrored Wasserstein loss in the discriminator to enforce better semantic-level reconstruction. We also propose a latent space regularization to learn a compact manifold for in-distribution samples. The use of AMA produces better feature representations that improve anomaly detection performance. Second, we put forward an alternative measure of anomaly score to replace the reconstruction-based metric which has been traditionally used in generative model-based anomaly detection methods. Our method outperforms the current state-of-the-art methods for anomaly detection on several OOD detection benchmarks.



\end{abstract}

\input{1_introduction}

\input{3_relatedwork}

\input{4.1_Approach}
\input{5.1_Experiments}

\input{9_Conclusion}

\section{Acknowledgements}
This project was supported in part by NSF CAREER AWARD 1942230, an IBM faculty award, a grant from Capital One, and a Simons Fellowship on Deep Learning Foundations. This work was supported through the IBM Global University Program Awards initiative. Authors thank Ritesh Soni, Steven Loscalzo, Bayan Bruss, Samuel Sharpe and Jason Wittenbach for helpful discussions.


{\small
\bibliographystyle{ieee_fullname}
\bibliography{egbib}
}

\end{document}

%% file: 1_introduction.tex
\section{Introduction}
\label{sec:introduction}
When deploying machine learning models in the real world, we need to ensure safety and reliability along with the performance. The models which perform well on the training data can be easily fooled when deployed in the wild \cite{nguyen2014deep,szegedy2013intriguing}. Recognizing novel or anomalous samples in the landscape of constantly changing data is considered an important problem in AI safety~\cite{amodei2016concrete}. Flagging anomalies is of utmost importance in many real-life applications of machine learning such as self-driving and medical diagnosis. The task of identifying such novel or anomalous samples has been formalized as Anomaly Detection (AD). This problem has been studied for years under various names, like novelty detection, out-of-distribution detection, open set recognition, uncertainty estimation, and so on~\cite{hodge2004survey, chandola2009anomaly, chalapathy2019deep}.

If the training data has the class labels available within the normal samples, several approaches have been proposed for OOD detection on top of or within a neural network classifier \cite{hendrycks2016baseline, liang2017enhancing, vyas2018out, hsu2020generalized, hendrycks2018deep, lee2018simple}. While these methods perform exceptionally well, they cannot be used in unsupervised or one class classification scenarios where labels are missing or not available for most of the classes, for instance, in credit card fraud recognition scenario, we are presented with lot of normal transactions, but no additional label available for transaction type. A rather obvious choice in such cases is to learn the underlying distribution of the data using generative models. Within deep generative models, two styles of approaches are popular, (1) Likelihood based models like Flow models or Autoregressive models, and use the estimated likelihood to recognise anomalies (2) AutoEncoder (AE) style approaches where reconstruction error of a given input is used to recognize the anomalies. While likelihood based approaches allow computation of exact likelihood for a given sample, they are found to assign high likelihood score to out-of-distribution samples as noted in the recent literature~\cite{choi2018waic, nalisnick2018deep, ren2019likelihood}. The goal of AE based approaches is to learn a good latent representation of data by either performing reconstruction, or adversarial training with a discriminator~\cite{schlegl2017unsupervised,zenati2018adversarially, akcay2018ganomaly, akccay2019skip, ngo2019fence}. In this work, we focus on the latter, i.e., the AE style methods and resolve two specific problems associated with them. 

First, the $\ell_{p}$ loss used for reconstruction by AutoEncoder (AE) methods compares only pixel-level errors but does not capture the high-level structure in the image. \cite{munjal2020implicit,rosca2017variational} proposed to alleviate this problem by introducing an adversarial loss~\cite{goodfellow2014generative}. While adversarial loss fixes the problem of blurry reconstructions in low-diversity settings such as CelebA~\cite{liu2018large} faces, quality of reconstruction remains poor for more diverse datasets such as CIFAR~\cite{krizhevsky2009learning} with many unrelated sub-classes like cats, and airplanes~\cite{munjal2020implicit}. We posit that this issue arises because the loss function in ~\cite{munjal2020implicit} compares distributions for a batch of samples but not the individual samples themselves. Hence a cat image reconstructed as an airplane is still a feasible solution since both airplane and cat belong to the same unlabeled input distribution. 
To address this problem, we propose Mirrored Wasserstein loss, where for a given sample $\bx$ and its reconstruction $\hbx$, a discriminator measures the Wasserstein distance between the joint distribution $(\bx,\bx)$ and $(\bx,\hbx)$. Stacking the image with its reconstruction allows discriminator to not only minimize the distance between distributions of images and reconstructions as before, but also ensures that each reconstruction is pushed closer to its ground truth. In \S~\ref{sec:Mirrored loss}, we give an intuition on how the mirrored Wasserstein loss improves the reconstructions quality as compared to the Wasserstein loss.

The second problem associated with AE methods is the regularization of latent space. In absence of explicit regularization, the model ends up overfitting the training distribution. Several regularization approaches have been proposed in the past~\cite{kingma2013auto,makhzani2015adversarial}, typically with a goal of sampling from the latent distribution. In our work, we consider regularizing the latent space of the model from the perspective of anomaly detection. Ideally, we want the latent space to be smooth and compact for the samples with in the distribution, while simultaneously pushing away out-of-distribution samples. To this end, we perform a simplex interpolation between latent representations of multiple samples in the training data, to ensure that decoder reconstructions of these latents are also realistic~\cite{berthelot2018understanding}. For the training purposes, we generate synthetic negative samples by sampling from atypical set in latent space~\cite{cover1999elements}. Our latent space regularizer ensures high quality reconstructions for in-distribution latent codes thus improving the Anomaly Detection performance as demonstrated quantitatively in Section~\ref{sec:experiments}.

In summary, our main contributions are:
\begin{itemize}
\item We propose \textbf{Adversarial~Mirrored~AutoEncoder (AMA)}, an AutoEncoder~Discriminator style network that uses Mirrored Wasserstein loss in the discriminator to enforce better reconstructions on diverse datasets.
\item We propose Latent space regularization during training by performing \textbf{Simplex Interpolation} of normal samples in the latent space and by sampling \textit{synthetic negatives} by \textbf{Atypical Selection} and optimizing the latent space to be away from them.
\item We propose an anomaly score metric that generates likelihood-like estimate for a given sample with respect to the distribution of reconstruction scores of training data.
\end{itemize}

We performed extensive evaluations on various benchmark image datasets CIFAR-10~\cite{krizhevsky2009learning}, CIFAR-100~\cite{krizhevsky2009learning}, ImageNet~(resized)\cite{deng2009imagenet}, SVHN~\cite{sermanet2012convolutional}, Fashion MNIST~\cite{xiao2017fashion}, MNIST~\cite{lecun1998gradient}, Omniglot~\cite{lake2019omniglot} and our model outperforms the current state-of-the-art generative methods for anomaly detection.

%% file: 3_relatedwork.tex
\section{Related work}
\label{sec:relatedwork}

\begin{figure*}[ht!]
\begin{center}
\includegraphics[width=0.9\linewidth]{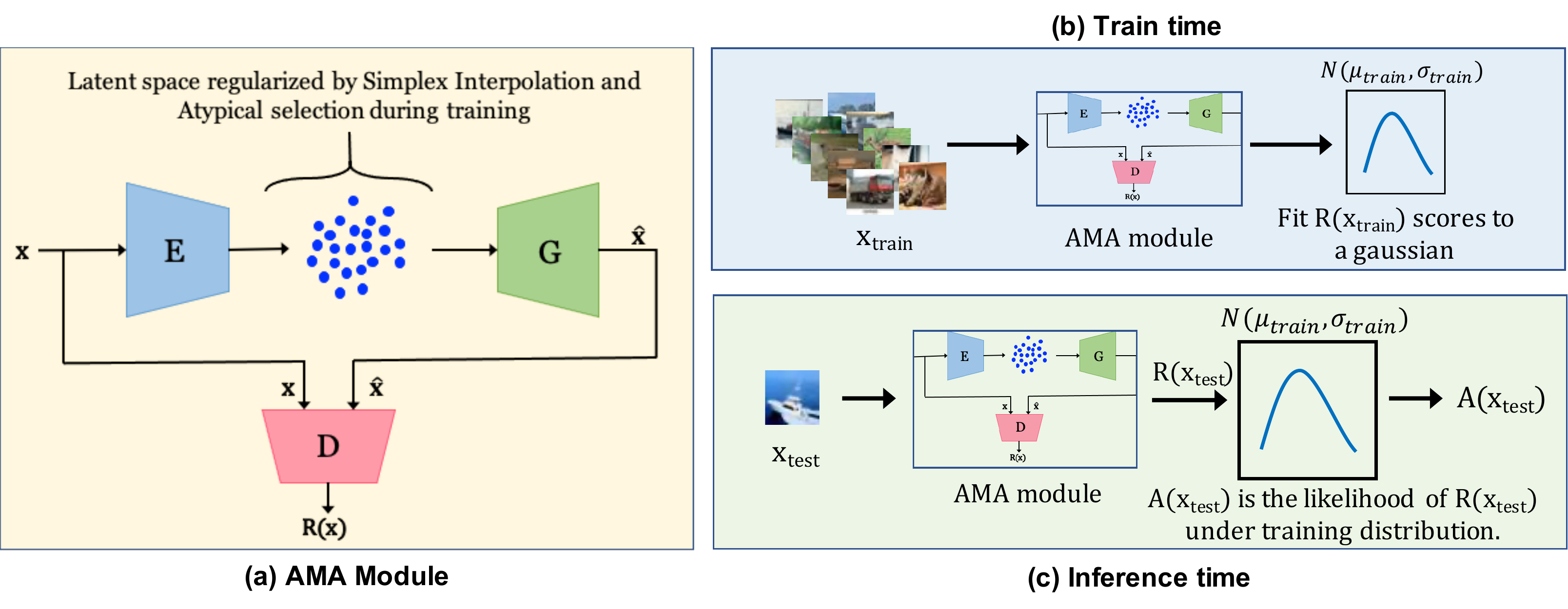}
\end{center}
  \caption{\textbf{AMA pipeline:} Our model consists of an Encoder $\bE$, a Generator $\bG$ and a Discriminator $\bD$. (a) First we train the model on all the training samples by optimizing the min-max objective from Eq.~\ref{eq:final_objective} with latent space regularization discussed in \S~\ref{sec:latent}. 
  (b) Next, we take the trained AMA module and pass the complete train data to generate R-scores using Eq.~\ref{eq:R_x} and fit them to a Gaussian distribution. (c) During inference, given an image $x_{test}$, we first calculate $R(x_{test})$ by passing it through frozen-AMA module, and then $A(x_{test})$ using Eq.~\ref{eq:A_x}, which is essentially the likelihood of $R(x_{test})$ under the Gaussian curve we generated in (b).  Lower the $A(x_{test})$, more the likelihood of the given test sample being anomalous.}
\label{fig:AMA_flow}
\end{figure*}

The problem we are trying to solve is OOD detection in datasets with no class labels. Depending on the field, it is studied under various names like One-class classification, Novelty detection, and so on. 

\noindent\textbf{Likelihood based approaches:}
Since generative modeling techniques such as Glow~\cite{kingma2018glow}, PixelRNN\cite{oord2016pixel}, or  PixelCNN++~\cite{salimans2017pixelcnn++} allow us to compute exact likelihood of data samples, several anomaly detection methods are built on the top of the likelihood estimates provided by these models. LLR~\cite{ren2019likelihood} proposes to train two models, one on the background statistics of the training data by random sampling of pixels and second model on the training data itself. Given an image, anomaly score is given by the ratio of likelihoods predicted by these two models. WAIC~\cite{choi2018waic} suggests to use Watanabe Akaike Information Criteria calculated over ensembles of generative model as anomaly scoring metric. Serra et al~\cite{serra2019input} proposes an $\mathcal{S}$-criterion, which is calculated by subtracting complexity estimate of the image from the negative log-likelihood predicted by a PixelCNN++ or a Glow model. Typicality test~\cite{nalisnick2019detecting} proposes a test for typicality of the samples by employing a Monte-carlo estimate of the empirical entropy. A limitation of this method is that it needs multiple images at the same time for evaluation.


Some recent studies ~\cite{choi2018waic, nalisnick2018deep, ren2019likelihood} suggest that deep generative models trained on a dataset (say CIFAR-10) assign higher likelihoods to some out-of-distribution (OOD) images (e.g. SVHN). This behaviour is persistent in a wide range of auto-regressive models such as Glow, PixelRNN, and PixelCNN++ and raises the question whether the likelihood provided by these approaches can be reliably used for detecting anomalies.

\smallskip
\noindent\textbf{AutoEncoders or GANs based methods:}
A number of methods proposed recently use a different kind of metric for scoring anomalies. In DeepSVDD~\cite{ruff2018deep}, an Encoder-Decoder network is used to learn the latent representations of the data while minimizing the volume of a lower-dimensional hypersphere that encloses them. They hypothesize that anomalous data is likely to fall outside the sphere, and normal data is likely to fall inside the sphere. This technique is inspired by traditional SVDD (Support Vector Data Description)~\cite{tax2004support} where a hypersphere is used to separate normal samples from anomalies. Ano-GAN~\cite{schlegl2017unsupervised} is one of the first works that uses Generative Adversarial Nets (GANs)~\cite{goodfellow2014generative} for anomaly detection. In this work, a GAN is trained only on normal samples. Since a GAN model is not invertible, an additional optimization is performed to find the closest latent representation for a given test sample. The anomaly score is computed as a combination of reconstruction loss and discriminator loss. FGAN~\cite{ngo2019fence} trains a GAN on the normal samples and uses a combination of adversarial loss and dispersion loss (distance based loss in latent space) to discover anomalies. \cite{akcay2018ganomaly,akccay2019skip} use a series of Encoder, Decoder and Discriminator networks to optimize the reconstructions as well as distance between the representations.  ALAD~\cite{zenati2018adversarially} uses BiGAN~\cite{donahue2016adversarial} to improve the latent representations of the data. Each of these methods use discriminator-based score for detecting anomalies.

A recent survey by \cite{chalapathy2019deep} does a comprehensive study of anomaly detection approaches.

\smallskip
\noindent\textbf{Negative Selection Algorithms (NSA):}
NSA is one of the early biologically inspired algorithms to solve one-class classification problem, first proposed by \cite{forrest1994self} to detect data manipulation caused by computer viruses. The core idea is to generate synthetic negative samples which do not match normal samples using a search algorithm and use them to train a downstream, supervised anomaly classifier ~\cite{dasgupta2002anomaly,coello2002approach,gonzalez2002combining}. Since the search space for negative samples for high dimension data can grow exponentially very large, it can be computationally very expensive to sample synthetic negatives~\cite{jinyin2011study}. Recent work by \cite{sipple2020interpretable} proposes a simpler approach to perform negative selection by using uniform sampling and building a binary classifier with positives and \textit{synthetic negatives} to perform anomaly detection task.


%% file: 4.1_Approach.tex
\section{Adversarial Mirrored AutoEncoder (AMA)}
\label{sec:Approach}

\begin{figure}[t]
\centering
\includegraphics[width=0.95\columnwidth]{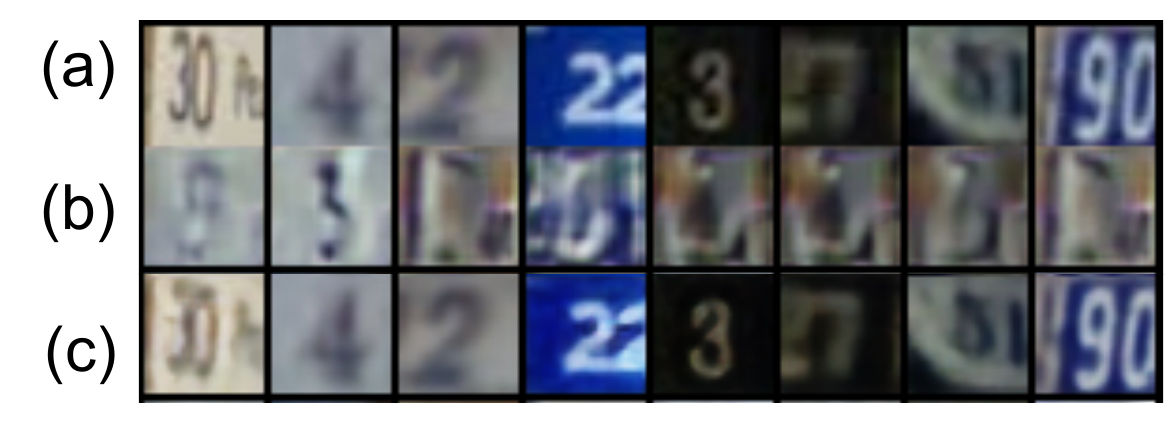} %
\caption{
Better reconstructions with Mirrored Wasserstein Loss. (a) Ground Truth (b) Reconstructions using AMA with regular Wasserstein loss (c) Reconstructions using AMA with Mirrored-Wasserstein loss. The quantitative comparisons are shown in Table~\ref{table:ood_aucs}}
\label{fig:semantic_recons}
\end{figure}

\begin{figure}[t]
\centering
\includegraphics[width=0.95\columnwidth]{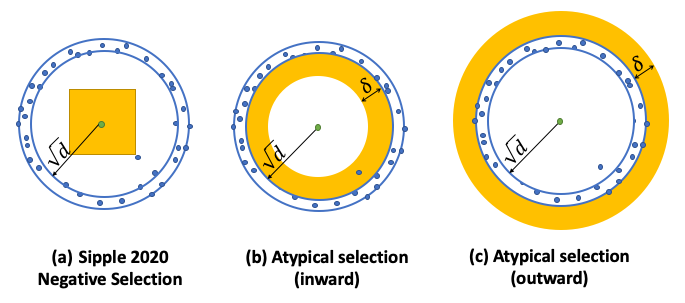}
\caption{An illustration of the negative sampling from atypical set in the latent space.
In each of the cases, typical set resides between the two blue d-dimensional spheres. Synthetic negative latents are drawn from yellow region (a) In \cite{sipple2020interpretable}, a cube centered at the origin is used as the negative sampling region. Instead, we propose to sample the synthetic negatives closer to the typical set between the spheres (b) $\sqrt{d}-\delta$ and $\sqrt{d}$  or (c) $\sqrt{d}$ and $\sqrt{d}+\delta$.}
\label{fig:atypical_selection}
\end{figure}

As discussed earlier, AMA consists of 2 major improvements over the conventional Auto-Encoder architectures: (i) Mirrored Wasserstein Loss, and (ii) Latent space regularization. These improvements help us outperform several state of the art likelihood, as well as reconstruction based anomaly detection methods. Fig.~\ref{fig:AMA_flow} shows an overview of our overall anomaly detection pipeline using AMA. In the following sub-sections, we  discuss each of the components of our anomaly detection framework in detail.

\subsection{Mirrored Wasserstein Loss}
\label{sec:Mirrored loss}
For training auto-encoders, $\ell_{1}$ or $\ell_{2}$ reconstruction loss between the original image and its reconstruction, defined as $\|\bx - \bx_{rec}\|_{p}$, is typically used. Reconstruction losses based on $\ell_{p}$ distances results in blurred decodings, thus producing poor generative models. Also, the use of $\ell_p$ reconstruction losses as anomaly scores, which is the standard technique used in Auto-Encoder based anomaly detection, has several limitations: (1) $\ell_{p}$ distances do not measure the perceptual similarity between images, which makes it hard to detect outliers that are semantically different, (2) A large $\ell_p$ reconstruction loss between input and its decoding can be an outcome of poor generative modeling and not because the image is an outlier. 

Motivated by the success of Generative Adversarial Networks (GANs) in obtaining improved generations, a number of approaches replace the $\ell_p$ reconstruction losses in Auto-Encoders with an adversarial loss that captures high-level details in the image. While this loss is good enough to get good reconstructions in low-diversity datasets like MNIST, CelebA, but it is not enough to reconstruct diverse datasets like CIFAR-10 or Imagenet \cite{munjal2020implicit}. 

Regular Wasserstein loss function only ensures the input and its generated sample both belong to the same distribution, but doesn't necessarily make input and its reconstruction look alike. 
To resolve this problem, for a given sample $\bx \sim \PP_{X}$ and its reconstruction $\hbx \sim \PP_{\hX} $, we perform a Wasserstein minimization between the joint distributions $\PP_{X, X}$ and $\PP_{X, \hX}$. The discriminator now takes in stacked pairs of input images $(\bx, \bx)$ and $(\bx, \hbx)$. This clearly avoids the problems discussed in the previous part as the distribution $(\bx, \bx)$ always has pairs of samples that are similar looking. If a car image is reconstructed as an airplane, the generated distribution will contain a (car, airplane) sample, which is never found in the input distribution $(\bx, \bx)$. Hence, the model will aim to generate samples sharing the same semantics. Figure~\ref{fig:semantic_recons} shows the difference in image reconstructions using AMA with regular Wasserstein loss \vs AMA with Mirrored Wasserstein loss. While both the models perform well in terms of image quality, we can see that for the first image, the ground truth is the number 30, and regular Wasserstein loss model is fitting number 9, though very unlike the ground truth, but still from the same distribution, while AMA with Mirrored Wasserstein loss is faithful to the ground truth and reconstructed a very similar looking 30.

Formally speaking, our model formulates a distribution of a set of samples  $\bx \sim \PP_{X}$, using the Mirrored Wasserstein loss, as follows:
\begin{align}\label{eq:wasserstein_joint}
&W(\PP_{X,X},\PP_{X,\hX}) = \max_{\bD \in Lip-1}~\mathop{\mathbb{E}}_{x\sim \PP_X}\left[{\bD(\bx,\bx) - \bD(\bx,\hbx)}\right]
\end{align} 
where $\hbx=\bG(\bE(\bx))$ and \emph{Lip-1} denotes the 1-Lipschitz constraint. Note that Eq.~\eqref{eq:wasserstein_joint} is similar to the loss function of Wasserstein GAN~\cite{martin2017wasserstein} with the only difference that discriminator $\bD$ acts on the stacked images $(\bx, \bx)$ and $(\bx, \hbx)$. This is equivalent to minimizing the 
Wasserstein distance between conditional distributions $W(\PP_{X|X}, \PP_{\hX|X})$. This model also shares similarities to discriminator architectures used in conditional image to image translations such as Pix2Pix~\cite{isola2017image}.

\begin{lemma}
If E and G are optimal encoder and generator networks, i.e., $\PP_{X,\bG(\bE(X))} = \PP_{X,X}$, then $\bx$ = $\bG(\bE(\bx))$.
\end{lemma}




\subsection{Latent Space Regularization}
\label{sec:latent}
The neural networks are universal approximators, and an autoencoder trained without any constraints on the latent space will tend to overfit the training dataset. While several regularization schemes have been proposed, in this section we develop our regularization framework adapted for the task for anomaly detection.

\medskip
\noindent\textbf{Simplex Interpolation in Latent space}

\smallskip
\noindent
\cite{berthelot2018understanding} showed that by forcing linear combination of a latent codes of a pair of data points to look realistic after decoding, the Encoder learns a better representation of data. This is demonstrated by improved performance on downstream tasks such as supervised learning and clustering. However, \cite{sainburg2018generative} argues that pairwise interpolation between samples of $\bx$ proposed by \cite{berthelot2018understanding} does not reach all points within the latent distribution, and may not necessarily make the latent distribution compact. Hence, we propose to use Simplex interpolation between $i$ randomly selected points to make the manifold smoother and amenable. 


Given $k$ normal samples $\bx_1, \bx_2,\dots, \bx_k$, we uniformly sample $k$ scalars $\alpha_i$ from $[0,0.5]$, we define an interpolated sample as: 
\begin{align*}
    \hat{\bx}_{inter} &= \bG\left(\frac{1}{\sum{\alpha_i}} (\alpha_1 \be_1 + \alpha_2 \be_2 + \dots \alpha_k \be_k ) \right) \\
    \be_i &= \bE(\bx_i) ~~~\forall i
\end{align*}
Here, $\hbx_{inter}$ denotes the interpolated latent point. A discriminator is then trained to distinguish between $(\bx, \bx)$ pair and $(\bx, \hat{\bx}_{inter})$ pair, while the generator learns by trying to fool the discriminator. That is,
\begin{align*}
\min_{\bG} \max_{\bD \in Lip-1}~\mathop{\mathbb{E}}_{x\sim \PP_X}\left[{\bD(\bx,\bx) - \bD(\bx,\hbx_{inter})}\right]
\end{align*} 
This ensures that the distribution of interpolated points follow the same distribution as the original data distribution, thereby improving the smoothness in the latent space. 
We use $k=3$ in all our experiments. We empirically observe that larger values of $k$ give marginal improvements.

\medskip
\noindent\textbf{Negative Sampling by Atypical Selection}

\noindent
In our experiments, we observed that regularization on the convex combination of latent codes of training samples works better if we also provide some negative examples, \ie, examples which should not look realistic. Since we are working in an unsupervised setting, we propose to generate synthetic negative samples in the by sampling from ``atypical set'' of the latent space distribution.

A typical set of a probability distribution is the set whose elements have information content close to that of the expected information. It is essentially the volume that not only covers most of mass of the distribution, but also reflects the properties of samples from the distribution. Due to the concentration of measure, a generative model will draw samples only from typical set~\cite{cover1999elements}. Even though the typical set has the highest mass, it might not have the highest probability density. Recent works \cite{choi2018waic, nalisnick2019detecting} propose that normal samples reside in typical set while anomalies reside outside of typical set, sometimes even in high probability density region. Hence we propose to sample outside the typical set in the latent space to generate synthetic negatives.

The Gaussian Annulus Theorem \cite{blum2016foundations, vershynin2018high} states that in a $d$-dimensional space, a typical set resides with high probability at a distance of $\sqrt{d}$ from the origin. In the absence of true negatives, we can obtain synthetic negatives by sampling the latents just outside and closer to the typical set than the origin and then use the generator for reconstruction. Although, our latent space is not inherently Gaussian, we observe that due to the $\ell_{2}$ regularization placed on the latent encodings, most of the training samples' encodings are close to $\sqrt{d}$ in magnitude. We sample atypical points uniformly between spheres with radii $\sqrt{d}$ and $\sqrt{d} \pm \delta$ as illustrated in Fig.~\ref{fig:atypical_selection}~(b)~(c). We call this procedure \textbf{Atypical Selection}. The $\delta$ and the direction of the selection, \textit{inward} or \textit{outward} are hyperparameters which are chosen based on the true anomaly samples available during the validation time. 

\cite{sipple2020interpretable} proposed a similar technique where \textit{synthetic negatives} are sampled around the origin as shown in Fig.~\ref{fig:atypical_selection}(a). We show in Table~\ref{table:atypicvssipple} that Atypical Selection outperforms this style of negative selection across multiple benchmarks.

\subsection{Overall objective}
Let $\mathbb{\widetilde{Q}}_{X}$ be the distribution of all atypical samples and let $\PP_X$ be the distribution of normal samples. We consider two different scenarios, first, when we don't have access to any anomalies during training, and the second case when we have access to a few anomalies.

\noindent\textbf{Unsupervised case:}
We train the AMA using the following min-max objective:

\begin{align}
   \min_{\bG}~~\max_{\bD \in Lip-1} & \cL_{normal} - \lambda_{neg} \cL_{neg}
\end{align}

 $\cL_{normal}$ part of the loss is to improve the reconstructions of normal in-distribution samples . It consists of 3 terms, first term is inspired by Mirrored Wasserstein loss, making sure that reconstructions look like their ground truths, second term is to ensure the interpolated points look similar to normal points, and the third term is a regularization term on encodings. $- \lambda_{neg} \cL_{neg}$ part is to penalize the anomalies. It ensures that anomalies are not reconstructed well. In this paper, since we assume that real anomalies are not available to us during training, we instead use generated \textit{synthetic anomalies} in this term. 

\begin{align}\label{eq:final_objective}
\cL_{normal} = &\mathbb{E}_{\bx \sim \PP_X} \Big[\bD(\bx,\bx) - \bD(\bx,\hbx) + \\
 & \lambda_{inter} \left( \bD(\bx,\bx) - \bD(\bx,\hbx_{inter}) \right) + \nonumber \\
 & \lambda_{reg} \|\bE(\bx)\| \Big] \nonumber \\
\cL_{neg} = &\mathbb{E}_{\bx \sim \mathbb{\widetilde{Q}}_{X}} \left[{\bD(\bx,\bx) - \bD(\bx,\hbx_{neg})}\right]
\end{align} 
 $\hbx_{neg} = G(\bz_{neg})$, where $\bz_{neg}$ is the latent sampled by Atypical Selection. $\lambda_{neg}$ is the Atypical Selection hyper-parameter, $\lambda_{inter}$ is the weight for the interpolation component, $\lambda_{reg}$ is the latent space regularization weight and $\|E(x)\|$ acts as regularizer for the latent representations.

\smallskip
\noindent\textbf{Semi-Supervised case:}
If we have a few true anomalies available during the training, we can use the same objective by using real anomalies instead of synthetic negatives in the $\cL_{neg}$ term. Please refer to appendix for related experiments. 

\begin{figure}[!ht]
\centering
\includegraphics[width=0.95\columnwidth]{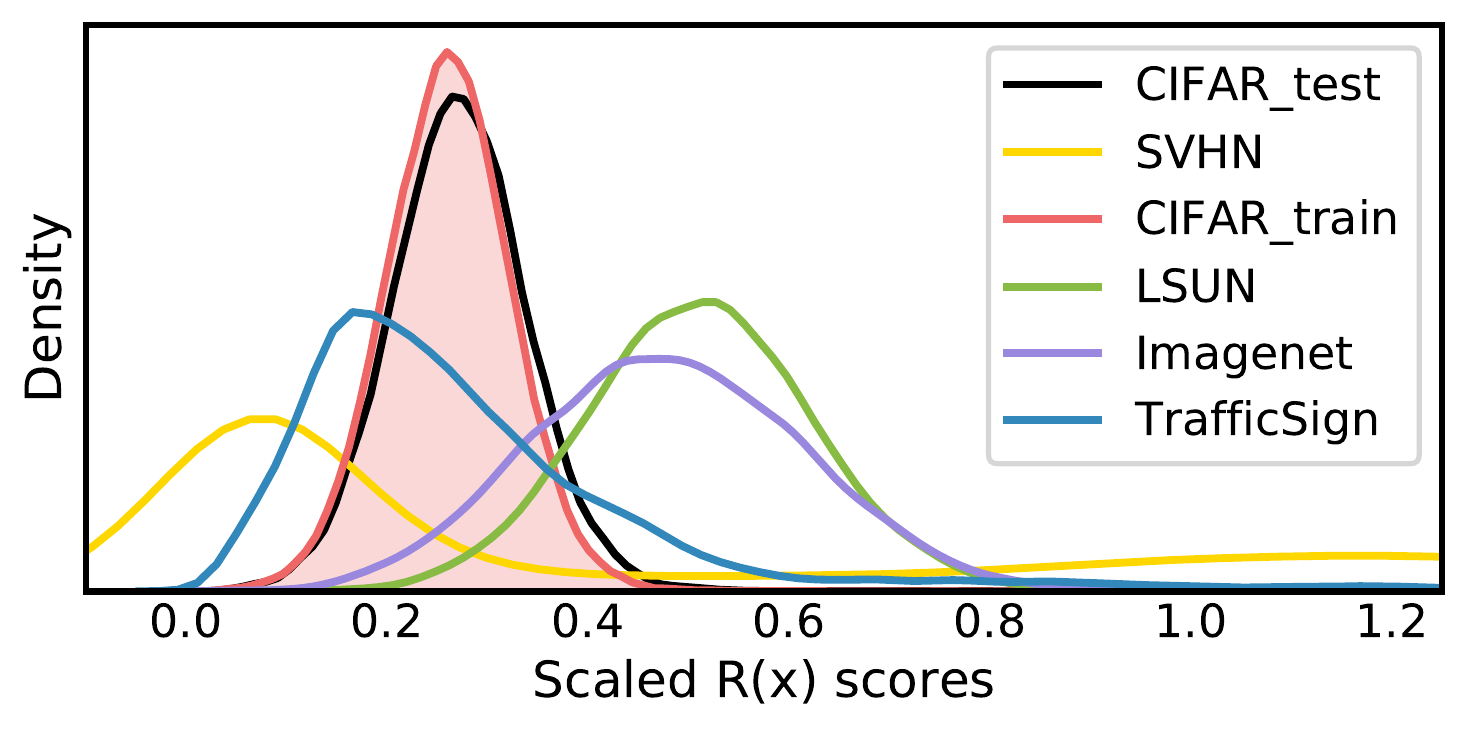} %
\caption{In this figure we show the density plots of R-scores predicted on various datasets by an AMA module trained on CIFAR-10 train data. We can note that CIFAR-10 train and test data distributions highly overlap. Depending on the dataset we are considering to be OOD distribution, R-scores of anomalous samples can trend lower or higher than the normal samples. If we use R-score directly to tag the anomalies, it will classify all the images with higher R-scores are anomalies, including many of the CIFAR-10 test samples. Meanwhile, all those to the left will be misidentified as normal samples. Instead of taking R-score on its face value, we propose to create A-score which weighs in the R-score of a given sample with respect to training data R-scores. R-score can be computed using Eq:\ref{eq:R_x} and A-score can be computed using Eq:\ref{eq:A_x}}
\label{fig:cifar_rx}
\end{figure}

\input{4.2_Anomalyscoring}

%% file: 4.2_Anomalyscoring.tex
\subsection{Anomaly score}
\label{sec:anomaly_metric}

Prior work in GAN-based anomaly detection used discriminator output as anomaly score~ \cite{schlegl2017unsupervised,ngo2019fence}. \cite{zenati2018adversarially} proposed an improvement by computing the distance between a sample and its reconstruction in the feature space of the discriminator, \textbf{R-score} (or R(x) score used interchangeably), which can be written as:
\begin{align}
\label{eq:R_x}
    R(x) = \|f(x,x) - f(x,\bG(\bE(x)))\|_1
\end{align}
where $f(\cdot,\cdot)$ is the penultimate layer of the discriminator. 

In \cite{zenati2018adversarially}, authors claim that the anomalous samples will have higher $R(x)$ values compared to that of normal samples. While this is true for the datasets considered in \cite{zenati2018adversarially}, we observed a counter-intuitive behaviour in some OOD detection scenarios.
In CIFAR-10 vs SVHN OOD detection experiment, our model and many other AE-based anomaly detectors~(including \cite{zenati2018adversarially}) assign lower R-scores to OOD samples as shown in Fig.~\ref{fig:cifar_rx}. 
This behavior is similar to the observations in \cite{nalisnick2018deep,choi2018waic} where sample likelihoods are used as anomaly scores. 
Even though the R-scores distribution of test CIFAR-10 samples overlaps with training distribution quite well, if we use the R-scores to compute AUROC, it results in a very low AUC value(0.442 from Table~\ref{table:ood_aucs}), meaning most of the anomalies are classified as normals. 
This suggests that this reconstruction-based score is not a robust anomaly scoring function in all OOD detection scenarios.

Hence we propose the following technique: (i) fit the R-scores of training data to a Gaussian distribution (ii) compute the anomaly score for a given test sample $x_i$ as the likelihood of $R(x_i)$ under the Gaussian distribution. The proposed anomaly metric, \textbf{A-score} (or A(x) score used interchangeably) can be written as:
\begin{align}
\label{eq:A_x}
    A(x_i) = \frac{1}{{\sigma \sqrt {2\pi } }}e^{{{ - \left( {R(x_i) - \mu } \right)^2 } \mathord{\left/ {\vphantom {{ - \left( {x - \mu } \right)^2 } {2\sigma ^2 }}} \right. \kern-\nulldelimiterspace} {2\sigma ^2 }}}
\end{align}
where $\mu$ is the mean and $\sigma^2$ is the variance of the distribution of R-scores over the training data.

A-score metric measures how similar the behaviour of test-time sample to that of training data, while R-score looks at relative behaviour of samples only at the test time. 

%% file: 5.1_Experiments.tex
\section{Experiments and Results}
\label{sec:experiments}
\begin{table*}
\centering
\caption{In this table we present AUROC scores on various OOD detection across various datasets. All methods in the table have no access to OOD data during training, but a small number of anomalies during validation to choose the best model. All the results are average AUROC values across test dataset, with one sample evaluated at a time except the result for Typicality test\cite{nalisnick2019detecting} which corresponds to using batchsize of 2 of the same type. In the bottom half of the table we show the ablation results of our model AMA with one component missing at a time from our pipeline.}
\resizebox{0.99\textwidth}{!}{
\begin{tabular}{lccccccccccc}
\hline
Trained on:                 & & \multicolumn{2}{c}{FashionMNIST}  & & \multicolumn{3}{c}{CIFAR-10} & &  \multicolumn{3}{c}{SVHN} \\
\cline{3-4}\cline{6-8}\cline{10-12}
OOD data:                            & & MNIST         & Omniglot      & & SVHN      & Imagenet    & CIFAR-100 && CIFAR-10      & Imagenet    & CIFAR-100   \\
\hline
WAIC on WGAN ensemble \cite{choi2018waic}                    & & 0.871         & 0.832        & & 0.623    & 0.626   & -  & & -     & -     & -     \\
Likelihood-ratio on PixelCNN++ \cite{ren2019likelihood}   & & \textbf{0.994}       & -        & & 0.931    &  -   & -  & & -     & -     & -   \\
Typicality test on Glow model \cite{nalisnick2019detecting}    & & 0.140         & -        & & 0.420    &  0.640   & -  & & 0.980     & \textbf{1.000}     & -   \\
DeepSVDD \cite{ruff2018deep}       & & 0.864        & 0.999           & & 0.533     & 0.387       & 0.478  & & 0.795     & 0.823    & 0.819      \\ 
$S$ using PixelCNN++ and FLIF \cite{serra2019input}              & & 0.967         & \textbf{1.000}       & & 0.929    &  0.589   & 0.535 & & -     & -     & -   \\ 
\midrule
AMA w/o Mirrored Wass. Loss (Ours)        & & 0.653         & 0.899       & & 0.800   & 0.526     & 0.510 & & 0.503     &  0.693   & 0.592    \\
AMA w/o Simplex Interpolation (Ours)        & & 0.960         & 0.998           & & 0.820    & 0.847     & 0.537 & & 0.991     & 0.993     & 0.987   \\
AMA w/o Atypical selection (Ours)        & & 0.894        & 0.997      & & 0.861   & 0.812    & 0.535 & & 0.990    & 0.991    & 0.987 \\
AMA w/o new anomaly scoring  (Ours)        & & 0.991        & 0.997       & & 0.442   & 0.890    & 0.501 & & \textbf{0.993}    & \textbf{1.000}     & \textbf{0.988 }  \\
AMA (Ours)        & & 0.987      & 0.998       & & \textbf{0.958}    & \textbf{0.911}    & \textbf{0.551} & & \textbf{0.993 }    & \textbf{1.000 }    & \textbf{0.988}   \\

\hline
\end{tabular}
}

\label{table:ood_aucs}
\end{table*}


\subsection{Experimental Setting}
 
\noindent\textbf{Datasets:} Following the setting in \cite{ren2019likelihood,choi2018waic,serra2019input,nalisnick2019detecting}, we use CIFAR-10, SVHN and FashionMNIST are taken as normal datasets. We evaluate the performance of the models when the anomalies are coming from each of the OOD datasets, ImageNet(resize), CIFAR-100, LSUN(resize), iSUN, CelebA,  MNIST, Omniglot, TrafficSign, Uniform random images, Gaussian random images.
We also consider the case when anomalies arise within the same data manifold (i.e. same dataset). We evaluated this scenario on CIFAR-10 and MNIST datasets. For these experiments, we consider one class as normal and rest of the 9 classes as anomalies following the setup from \cite{ruff2018deep, zenati2018adversarially}.

\smallskip
\noindent\textbf{Baselines:} We compare our model against various generative model based anomaly detection approaches. Ren et al \cite{ren2019likelihood} uses likelihood based estimate from a Autoregressive model to discover anomalies. WAIC~\cite{choi2018waic} proposes to use WAIC criteria on top of likelihood estimation methods to find anomalies. Serra et al \cite{serra2019input} leverages complexity estimate of images to detect OOD inputs. Typicality test~\cite{nalisnick2019detecting} proposes to calculate an empirical estimate of entropy of set of samples and use it to recognize anomalies. DeepSVDD \cite{ruff2018deep} optimizes the latent representations of images and uses the distance in latent space as complexity measure. 

In addition to these, another set of methods \cite{akccay2019skip,zenati2018adversarially, schlegl2017unsupervised, ngo2019fence, ruff2018deep} addresses the scenario of anomalies from the same data manifold (i.e. same dataset) in their respective papers. We have compared our model to these methods in this scenario as well and we believe these methods are just as applicable to OOD samples coming from different data manifold. For these experiments we follow the setup from \cite{zenati2018adversarially, ruff2018deep, schlegl2017unsupervised}, where one class is considered normal and the rest of the classes from the same dataset as anomalies. All the results shown in Table~\ref{table:indistribution_aucs} are for this setting. In DeepSVDD, Global Contrast Normalization is used on the data prior to the training. We removed this additional normalization step to make the method comparable to other baselines.

Note that discriminative models such as \cite{hendrycks2016baseline,hendrycks2018deep,hsu2020generalized} achieve high performance in several OOD detection benchmarks, but assume access to the class labels during training. For brevity, we consider only unsupervised baselines in this work.

\smallskip
\noindent\textbf{BatchNorm Issue:} While we were working on the baselines, we noticed that one of the earlier work~\cite{akccay2019skip}\footnote{https://github.com/samet-akcay/skip-ganomaly} has evaluated their model in the training mode instead of setting it in the evaluation mode. Due to this issue, the BatchNorm is calculated for the test batch, rather than using the train-time statistics. 
Hence, while reporting results for~\cite{akccay2019skip}, we re-evaluate their models by freezing the BatchNorm statistics during the test time. We follow the same protocol in the case of rest of the models as well.

\smallskip
\noindent\textbf{Network Architectures and Training:}
The generator and the discriminator architectures have residual architectures and are borrowed from Spectral Normalization GAN~\cite{miyato2018spectral}. Our Encoder is a 4 layered convolution network with BatchNorm and LeakyRelu nonlinearity. Refer to appendix for the complete architecture details.

Following the setting in \cite{zenati2018adversarially, ren2019likelihood} we assume that we have access to a small number of anomalies during validation time ($\approx 50$ in number). To generate the test set, we randomly sample anomalies from the an OOD dataset, 20\% the size of normal samples, compared to sampling equal number of normal and anomalies scenario presented in \cite{ren2019likelihood,choi2018waic}. We believe our scenario is far more realistic and more stringent.We keep the test data and normalizations same for our model as well as the baselines to make them comparable.

The whole pipe-line of our model, AMA, is trained end-to-end with Adam optimizer with $\beta_1 = 0$ and $\beta_2=0.9$ for Generator and Discriminator and $\beta_1 = 0.5$ and $\beta_2=0.9$ for Encoder, initial learning rate of 3e-4 and decaying it by a factor of 0.1 at 30, 60 and 90 training epochs. We trained each model for 100 epochs with a batch size of 256 for all the datasets. If Atypical selection is enabled, we train the model for first the 10 epochs only on normal samples, and from $11^{th}$ epoch onwards we generate synthetic anomalies and use them along with normal samples in training. We use $\lambda_{inter} = 0.5$,  $\lambda_{neg}=5$ and $\lambda_{reg}=1$ in all of our CIFAR-10 and SVHN experiments. Refer to the appendix for the hyperparameter values of MNIST experiments. In OOD experiments, for CIFAR-10, we sampled for synthetic anomalies \textit{inward} and for SVHN and Fashion MNIST we sampled \textit{outward}. Experiments are performed using two NVIDIA GTX-2080TI GPUs.

\input{5.2_Results}

%% file: 5.2_Results.tex
\subsection{Anomaly Detection performance}
\label{sec:results}


\begin{table*}
\centering
\caption{Here we show performance of anomaly detection task when anomalies come from an unseen class of the same dataset. Each column denotes the normal class and the rest 9 classes from that respective dataset are considered as anomalies. The performance is measured using AUROC scores, higher the better.}
\resizebox{0.85\textwidth}{!}{
\begin{tabular}{l | c c c c c c c c c c | c } 
 \toprule
 MNIST & 0 & 1 & 2 & 3 & 4 & 5 & 6 & 7 & 8 & 9 & Average \\
 \midrule
 
  FGAN\cite{ngo2019fence} & 0.754 & 0.307 & 0.628 & 0.566 & 0.390 & 0.490 & 0.538 & 0.313 & 0.645 & 0.408 & 0.504  \\ 
 ALAD\cite{zenati2018adversarially} & 0.962 & 0.915 & 0.794 & 0.821 & 0.702 & 0.79 & 0.843 & 0.865 & 0.771 & 0.821  & 0.828 \\ 
 Ano-GAN\cite{schlegl2017unsupervised} & 0.902 & 0.869 & 0.623 & 0.785 & 0.827 & 0.362 & 0.758 & 0.789 & 0.672 & 0.720  & 0.731 \\ 
 Skip-Ganomaly\cite{akcay2018ganomaly} & 0.297 & 0.877 & 0.393 & 0.486 & 0.618 & 0.540 & 0.455 & 0.633 & 0.426 & 0.584 & 0.531  \\ 
 DeepSVDD\cite{ruff2018deep} & 0.971 & 0.995 & 0.809 & 0.884 & \textbf{0.920} & 0.869 & 0.978 & 0.940 & \textbf{0.900} & 0.946  & 0.921  \\ 
 AMA (Ours) & \textbf{0.986} & \textbf{0.998} & \textbf{0.882} & \textbf{0.891} & 0.894 & \textbf{0.938} & \textbf{0.981} & \textbf{0.983} & 0.876 & \textbf{0.948}  & \textbf{0.938}      \\   [1ex] 
 \midrule
 CIFAR-10 & airplane & automobile & bird & cat & deer & dog & frog & horse & ship & truck & Average \\ [0.5ex] 
 \midrule
  FGAN\cite{ngo2019fence} & 0.572 & 0.582 & 0.505 & 0.544 & 0.534 & 0.535 & 0.528 & 0.537 & 0.664 & 0.338 & 0.567  \\ 
 ALAD\cite{zenati2018adversarially} & 0.679 & 0.397 & 0.685 & \textbf{0.652} & 0.696 & 0.550 & 0.704 & 0.463 & \textbf{0.787}  & 0.391  & 0.601  \\ 
 Ano-GAN\cite{schlegl2017unsupervised} & 0.602 & 0.439 & 0.637 & 0.594 & \textbf{0.755} & 0.604 & \textbf{0.730} &  0.498 & 0.675 & 0.445 & 0.598  \\ 
 Skip-Ganomaly\cite{akccay2019skip} & 0.655 & 0.406 & 0.663 & 0.598 & 0.739 & 0.617 & 0.638 & 0.519 & 0.746 & 0.387 & 0.597   \\ 
 Deep SVDD\cite{ruff2018deep} & 0.682 & 0.477 & 0.679 & 0.573 & 0.752 & 0.628 & 0.710 & 0.511 & 0.733 & 0.567 & 0.631  \\ 
 AMA (Ours) & \textbf{0.752} & \textbf{0.634}  & \textbf{0.696} & 0.603 & 0.733 & \textbf{0.650} & 0.658  & \textbf{0.582} & 0.754 & \textbf{0.632} & \textbf{0.669} \\ 
\bottomrule
\end{tabular}
}

\label{table:indistribution_aucs}
\end{table*}






We consider two common scenarios used in literature to benchmark the performance of Anomaly Detection techniques. In the first scenario, we consider images from a given dataset as the normal samples and images from a different dataset (typically with a different underlying distribution) as anomalies. In the second scenario, we consider images from one of the categories in the dataset as normal images while all other as anomalies. Note that, in some papers, these two scenarios are referred as as out-of-distribution (OOD) and in-distribution anomalies. We do not make this distinction and use the term ``anomalies" to refer to the either scenario.

\medskip
\noindent\textbf{Images from different dataset as anomalies}
In Table~\ref{table:ood_aucs}, we show the performance of our model and the baselines against 3 different cases. Our first set of experiments uses gray-scale images from Fashion MNIST as normal images while the images from MNIST and Omniglot as OOD images. This is a relatively simple scenario and nearly all the baselines and our model achieve almost perfect AUROC. Even though our model does not have the best AUROC, it is well within the margin of error of the best performing the model. 

Next two cases are a bit more challenging when the images are colored and more diverse. In first case, we use normal samples from CIFAR-10, and anomalies from SVHN, Imagenet, and CIFAR-100. In the second case, we use normal samples from SVHN, while the anomalies coming from CIFAR-10, Imagenet and CIFAR-100.
Our model outperforms all the baselines in both these experiments.
This shows that our model, AMA is optimizing the latent space of normal samples well which leads to an impressive generalization behavior. Even though AUROC scores are greater than 0.9 in most of the cases, our model falls short in case of CIFAR-10 vs CIFAR-100 (similar behavior is observed for the other baselines as well). This is a really hard scenario and even humans will have tough time deciding whether a given image is from CIFAR-10 or CIFAR-100. 

\medskip
\noindent\textbf{Images from different categories as anomalies}
In Table-\ref{table:indistribution_aucs}, we show Anomaly Detection experiments when anomalies arise from the same data manifold (i.e. same dataset). Each column shows the results of a normal class with the rest of 9 classes as anomalies. Our method~(AMA) outperforms other methods in terms of average scores with 1.7\% AUROC gain over the next best method on MNIST and 3.7\% gain on CIFAR-10 dataset. In terms of an individual case comparison, we best 8 out of 10 cases on MNIST, while 6 out of 10 cases on CIFAR-10.

\subsection{Ablation studies}

We have introduced 3 main ideas in this paper: Mirrored Wasserstein loss, Latent space regularization using Simplex Interpolation and Atypical Selection, and an alternative Anomaly scoring technique. In the second half of the Table~\ref{table:ood_aucs}, we show the ablation results, removing one component at a time. As expected, removing Mirrored Wasserstein loss reduces the AUROC scores the most. AUROC scores are reduced by an order of $\sim 0.1$ points whenever a part of Latent space regularization is removed. We see that in most of the cases, removing Atypical Selection reduces the scores a bit more than removing Simplex Interpolation. The new anomaly scoring metric contrtibutes the most when the normal sample distribution is more diverse than the OOD distribution, eg: the case of CIFAR-10 as normal and SVHN as OOD. When we used R-score to identify anomalies in this scenario, most of the SVHN samples are tagged normal while most of the CIFAR-10 images tagged as anomalies, thus resulting in lower AUROC. 


\smallskip
\noindent \textbf{Atypical selection vs Sipple 2020:} Performance comparision of Atypical Selection against Negative Sampling proposed in \cite{sipple2020interpretable} is presented in Table~\ref{table:atypicvssipple}. Atypical Selection outperforms \cite{sipple2020interpretable}'s technique in all studied cases. We hypothesize that, since Atypical Selection samples near the boundary of the latent space, it enforces the encoder to create more compact latent space for normal samples.
\smallskip
\begin{table}[h!]
\centering
\caption{In this table we present the performace of various models trained using Negative selection proposed by Sipple~\cite{sipple2020interpretable} \textit{vs} Atypical Selection proposed by us, in presence and absence of Simplex interpolation. Values reported are AUROC scores in format Sipple / Atypical Selection }
\resizebox{1\columnwidth}{!}{
\begin{tabular}{ c c c c} 
 \toprule
   Experiment & No interpolation & With interpolation \\ 
 \midrule
   FashionMNIST \textit{vs} MNIST & 0.778 /\textbf{0.960} & 0.824 / \textbf{0.987} \\
    CIFAR-10 \textit{vs} SVHN & 0.752 / \textbf{0.820}  & 0.819 / \textbf{0.958} \\
    SVHN \textit{vs} CIFAR-10 & 0.723/\textbf{0.991} & 0.896/\textbf{0.993} \\
    
 \bottomrule
\end{tabular}
}

\label{table:atypicvssipple}
\end{table}

%% file: 9_Conclusion.tex
\section{Conclusion}
\label{sec:conclusion}

In this paper, we have  we introduced a new method for the unsupervised anomaly detection problem, Adversarial Mirrored Autoencoder~(AMA), equipped with Mirrored Wasserstein loss and a latent space regularizer. Our method outperforms existing generative model based anomaly detectors on several benchmark tasks. We also show how each of the components contribute to the model's performance in diverse data settings. While our model is quite powerful in OOD detection, it still underperforms in some data settings like CIFAR-10 vs CIFAR-100. This is rather similar to the setting of anomalies arising from the same data manifold. While we showed some early results in Table~\ref{table:indistribution_aucs}, we can further extend this work to improve for such scenarios.